%% file: main.tex
\documentclass[letterpaper, 10pt, conference]{ieeeconf}  

\IEEEoverridecommandlockouts                              
\usepackage[utf8]{inputenc}
\usepackage[T1]{fontenc}
\usepackage{graphicx}
\usepackage{listings}
\usepackage{breqn}
\usepackage{booktabs}
\usepackage{pythontex}
\usepackage{comment}
\usepackage{tikz}
\usetikzlibrary{tikzmark}
\usepackage{diagbox}
\usepackage{slashbox}
\graphicspath{ {./} }

\title{\LARGE \bf
Active Information Gathering for Long-Horizon Navigation Under Uncertainty by Learning the Value of Information
}

\author{Raihan Islam Arnob and Gregory J. Stein\thanks{R. Arnob and G. Stein are with the Department of Computer Science, George Mason University, USA, \{\texttt{rarnob}, \texttt{gjstein}\}\texttt{@gmu.edu}}}

\begin{document}
\maketitle
\thispagestyle{empty}
\pagestyle{empty}

\input{0_abstract}
\input{1_introduction}
\input{7_related_works}
\input{2_problem_formulation}

\input{4_example_case.tex}

\input{5_methodology}
\input{6_results}

\input{8_conclusion_future_work.tex}

\input{9_acknowledgement.tex}
\bibliographystyle{IEEEtran}
\input{main.bbl}


\end{document}

%% file: 0_abstract.tex
\begin{abstract}
We address the task of long-horizon navigation in partially mapped environments for which active gathering of information about faraway unseen space is essential for good behavior. 
We present a novel planning strategy that, at training time, affords tractable computation of the value of information associated with revealing potentially informative regions of unseen space, data used to train a graph neural network to predict the goodness of temporally-extended exploratory actions. Our learning-augmented model-based planning approach predicts the expected value of information of revealing unseen space and is capable of using these predictions to actively seek information and so improve long-horizon navigation. Across two simulated office-like environments, our planner outperforms competitive learned and non-learned baseline navigation strategies, achieving improvements of up to 63.76\% and 36.68\%, demonstrating its capacity to actively seek performance-critical information.
\end{abstract}

%% file: 1_introduction.tex
\section{Introduction}
We focus on the task of point-goal navigation in partially mapped large building scale environments.
In particular, we focus on environments in which active gathering of information is essential to reach the goal quickly: i.e., in minimum expected cost in terms of distance traveled. 
Consider the example of a human walking through a building that they have never been in before. 
The human looking for a particular location in unseen space is uncertain about how best to get there, since they do not know the surrounding building layout. 
If they are sufficiently uncertain about where to go next, it may be valuable to seek out a map of the building, often located in the security offices or near the elevator bays.

\begin{figure}[t]
  \includegraphics[width=.48\textwidth]{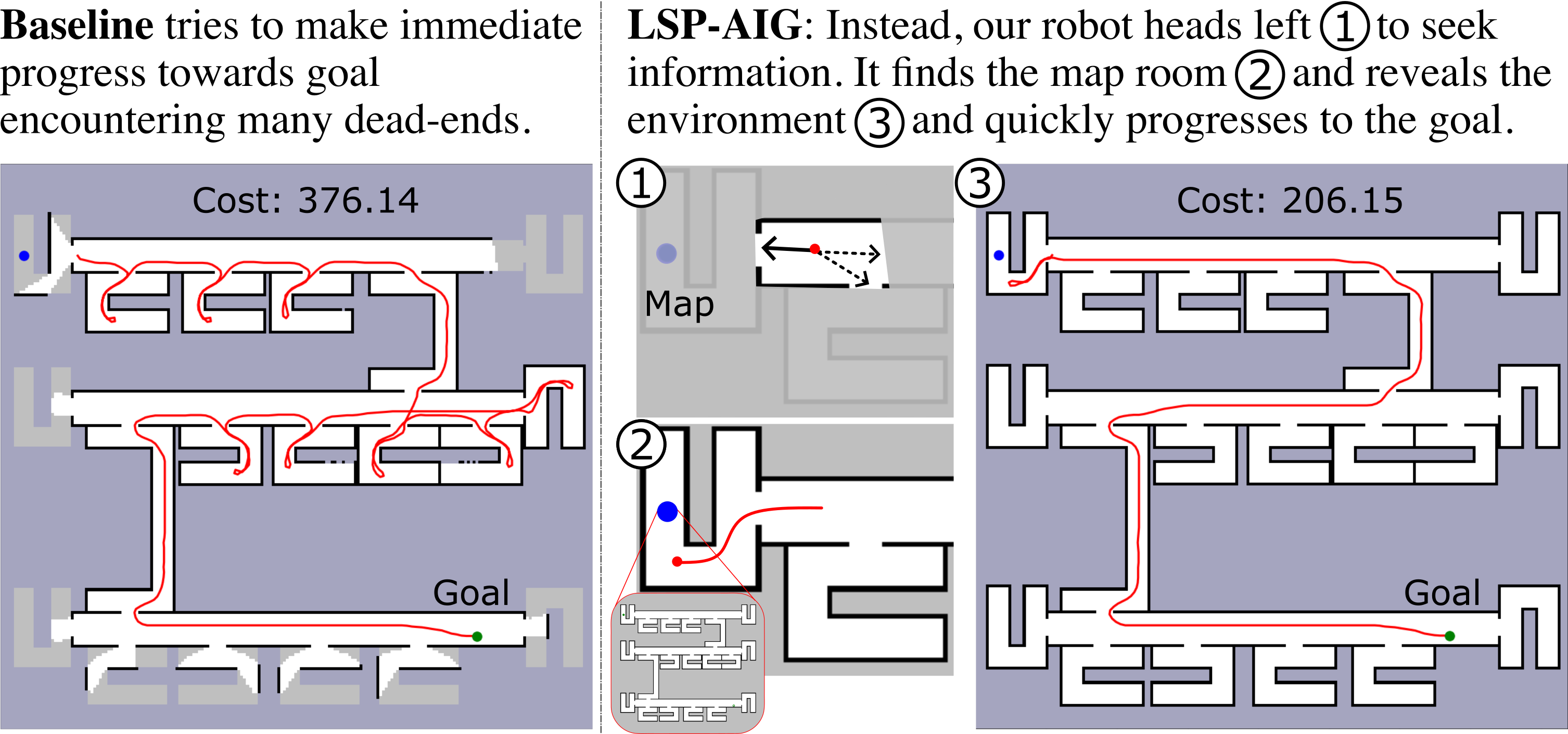}
  \vspace{-5pt}
  \caption{
  \textbf{Overview: actively gathering information is essential for good navigation in a partial map.} 
  Baseline approach reaches the goal slowly, encountering many dead-ends, while our approach (LSP-AIG) actively gathers useful information that lets it quickly reach the goal.
  }
  \vspace{-10pt}
  \label{fig:intro-explain}
\end{figure}

We show this information-seeking behavior in the scenario in Fig.~\ref{fig:intro-explain}, where a robot with a partial map aims to reach a faraway point goal in a large unmapped office building. 
An uninformed approach (left) often heads right, following the hallway in greedy pursuit of the goal; in the absence of additional information, this strategy encounters many dead-end rooms as it searches for a passage through to the next hallway needed to reach the unseen goal. 
Instead, a more effective strategy in this environment is to go out of the way to the end of the hallway in search of valuable information: in this case a map of the environment, found in a \emph{map room} near the start.
Once the map is found, the robot can use it to quickly reach the goal.
This map is valuable because of the information it provides. 
Thus, good behavior in general involves seeking out the map (or other navigation-relevant information) when the cost-savings that information is expected to provide outweighs the time the robot expects to spend locating or reaching the source of information.

It is the aim of this work to imbue a robot with the general purpose ability \textbf{to estimate the long-horizon value of information} associated with revealing regions of unseen space and to use those estimates \textbf{to encourage information seeking behavior when expected to improve performance}, affording both high-performance navigation with a sound and complete planning in large partially-mapped environments.

In general, incentivizing information gathering via classical approaches requires belief space planning, in which the robot envisions/imagines all possible observations and its subsequent behavior.
But belief based planning is canonically doubly exponential in time~\cite{Pineau-2002-8519} and so reasoning about long-horizon information gathering is intractable in general. 
However, considering the importance of information gathering actions for effective navigation in large-scale environments, we want to enable that ability in our robot.

Many approaches rely on learning for planning under uncertainty to mitigate the computational complexity of belief-space planning.
Learning-driven approaches---including many model-free approaches trained via deep reinforcement learning~\cite{MERLIN2018_Greg_Wayne, mirowski2018,dd-ppo}---have demonstrated the capacity to perform well in navigational planning under uncertainty and can exhibit information seeking behavior to complete their objectives. 
However, in the absence of an explicit map to keep track of where the robot has yet to explore, many such approaches are incomplete, lacking guarantees that they will reach the goal~\cite{pfeiffer2016}, and can perform poorly in large environments.

The recent Learning over Subgoals Planning (LSP) abstraction~\cite{pmlr-v87-stein18a,arnob2023lspgnn} uses a learning-augmented model-based strategy that has proven effective for performant and reliable planning under uncertainty. 
The approach uses \emph{frontiers}, boundaries between free and unseen space, to define temporally-extended exploratory actions and leverages learning to help determine the goodness of each. 
Despite the advantages of the LSP approach, it is not straightforward to determine how much value revealing a region of unseen space will provide and thus how the expected value of information provided by an exploratory action might be estimated during deployment and incorporated into planning.

We present a novel general purpose approach for long-horizon point-goal navigation in a partial map that (i) allows us to compute the value of information for an exploratory action---the improvement to the robots plan if some part of the environment were revealed---at training time, that (ii) can estimate and use value of information at planning time to encourage information seeking behavior when appropriate to improve plan performance, and (iii) is complete and sound. 
We show the effectiveness of our approach in procedural, simulated office-like environments, demonstrating improvements in average cost of up to 63.76\% and 36.68\% over non-learned and learned LSP-based baselines respectively, and reaches the unseen goal in 100\% of the trials.

%% file: 7_related_works.tex
\section{Related Works}\label{sec:related-works}
\textbf{Planning under Uncertainty}
POMDPs~\cite{kaelbling1998, littman1997, thrun2005,parascandolo-DCMCTS} have been used to represent navigation and exploration tasks under uncertainty, yet direct solution of the model implicit in the POMDP is computationally infeasible.
To mitigate this difficulty, many approaches to planning rely on learning to inform behavior~\cite{pfeiffer2016, richter2014, ross2013learning}, yet many~\cite{richter2014, thesis2017Richter} only plan a few time steps into the future and so are not well-suited to long-horizon navigation.
Some reinforcement learning approaches that target partially observed environments~\cite{DuanSCBSA16, yang2021,gupta2017cognitive, zhang2017deep, tai2017virtual, MirowskiPVSBBDG16} are also limited to planning in fairly small-scale environments.
The MERLIN agent~\cite{MERLIN2018_Greg_Wayne} uses a differentiable neural computer to recall information over much longer time horizons than is typically possible for end-to-end-trained model-free deep reinforcement learning systems.
The DD-PPO~\cite{dd-ppo} approach uses over 2.5 billions of images to train an end-to-end learning agent for indoor navigation under uncertainty.
However, such approaches~\cite{MERLIN2018_Greg_Wayne,Kober2014, henderson2018deep} can be difficult to train and lack plan completeness, making them somewhat brittle in practice.

\textbf{Information Gathering}
Most approaches that explicitly aim to encourage information gathering behavior focus on relatively short-time-horizon objectives---such as maximizing exploration~\cite{lodel2022look, Velez2011}, or helping to improve progress towards a vision-language navigation objective~\cite{wang2020active}, or to reveal uncertain object properties---or otherwise seek only to reveal the environment, maximizing coverage within a time budget~\cite{choudhury2017learning,multi2018Brent,li2023modelexplore,ramakrishnan2021exploration}.
Another approach from Zhang et al.~\cite{zhang2023embodied} focuses on very localized uncertainty: i.e., uncertain object properties and taking actions to reveal those uncertain properties, trading off between information gathering (about that particular object) and the cost of doing so.
Our proposed work improves long-horizon navigation under uncertainty by estimating long-horizon value of information for unseen spaces and actively gathering the information that is expected to improve performance.

%% file: 2_problem_formulation.tex
\section{Problem Formulation}
\label{sec:prob-form}
Our robot is tasked to find an unseen goal in a partially-mapped environment in minimum expected cost (distance).
The simulated robot is equipped with a semantically-aware planar laser scanner, which it can use to both localize and update its partial semantic-occupancy-grid map of its local surroundings, limited by range and obstacle occlusion.
As the robot navigates the partially-mapped environment, it updates its belief state $b_t$ to include newly-revealed space and its semantic class.
Our belief state $b_t$ can be represented as a two-element tuple consisting of the partially observed map $m_t$ and the robot pose $q_t$ : $b_t = \{m_t , q_t\}$.
At each time step $t$, the action $a_t$ specifies which of a set of dynamically-feasible motion primitives the robot executes.

Formally, we represent this problem as a Partially Observable Markov Decision Process~\cite{kaelbling1998,littman1997} (POMDP). The expected cost $Q$ under this model can be written via a belief space variant of the Bellman equation~\cite{Pineau-2002-8519}:
\begin{equation}
\label{eq:POMDP}
\begin{split}
    Q(b_t,a_t) = \sum_{b_{t+1}} P(b_{t+1}|b_t,a_t)\Big[R(b_{t+1},b_t,a_t) \\[-5pt]
    \quad\quad\quad + \min_{a_{t+1} \in \mathcal{A}(b_t+1)}Q(b_{t+1},a_{t+1})\Big],
\end{split}
\end{equation}
where $R(b_{t+1}, b_t, a_t)$ is the cost to reach belief $b_{t+1}$ from $b_t$ via action $a_t$ and $P(b_{t+1}|b_t, a_t)$ is the transition probability.

%% file: 4_example_case.tex
\section{Motivating Example: Illustrating the Importance of Active Information Gathering}
\label{sec:example-case}
\begin{figure}[t] 
  \includegraphics[width=.48\textwidth]{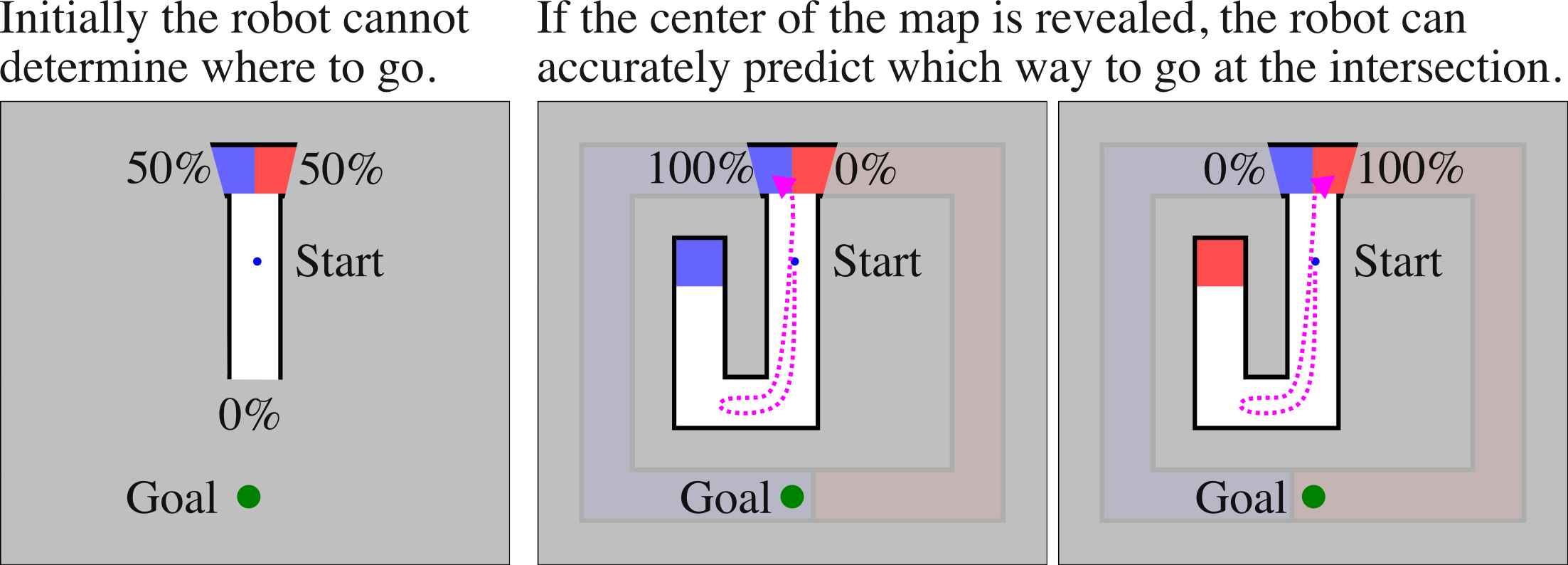}
    \vspace{-5pt}
    \caption{
    \textbf{Low cost navigation in our J-Intersection environment requires active gathering of information.}
    When the goal is either on left or right from the intersection, 
    knowing the information contained at the center of the map allows us to decide correctly at the intersection. 
    Choosing always left or right or even choosing one color over another will not reliably succeed.
    }
  \vspace{-10pt}
  \label{fig:example-case-j-shaped}
\end{figure}

We present a motivating example where active information gathering is needed to make good predictions about the environment while trying to reach an unseen goal.
Our \emph{J-Intersection} environment, as shown in Fig.~\ref{fig:example-case-j-shaped}, has either a red or blue square region inside of it and around the corner occluded from that square region far away at the intersection that colored region leads to the goal (bottom).
The environment is structured so that the color of the hallway the robot should follow matches the color at the center.

In this setting, it is impossible to make an informed decision at the intersection without the information that lies around the corner at the center.
Thus, optimal behavior in this environment requires that the robot go out of its way and explore downwards to reveal the center of the map, accumulating a short-term cost of exploration so that it may later benefit from the knowledge it gains from doing so.

For this motivating example, it is straightforward to determine how much value revealing the center region of the map provides.
More generally, quantifying the value of revealing a region of unseen space is incredibly difficult; doing so may require computing how the robot's behavior would change over the course of a lengthy deployment for all possible configurations of unseen space, motivating the need for a novel approach that can tractably quantify or estimate the value of information.

%% file: 5_methodology.tex
\section{Approach: Computing, Estimating, and Using Long-Horizon Value of Information to Navigate}\label{sec:eq-theory-value-of-info}

We present an approach to navigation under uncertainty that allows a robot to actively gather information specifically useful in improving its planning performance. 
This section describes (i) the learning-augmented model-based \emph{Learning over Subgoals} state and action abstraction of Stein et al.~\cite{pmlr-v87-stein18a} upon which our approach relies (Sec.~\ref{sec:lsp}), (ii) our additions to this abstraction to encourage information-seeking behavior (Sec.~\ref{sec:plan-with-value-of-info}), and (iii) our novel methodology for computing this value of information at training time (Sec.~\ref{sec:compute-value-of-info}) so that the expected value of revealing unseen space can be estimated during deployment.

\begin{figure}[t]
  \includegraphics[width=.48\textwidth]{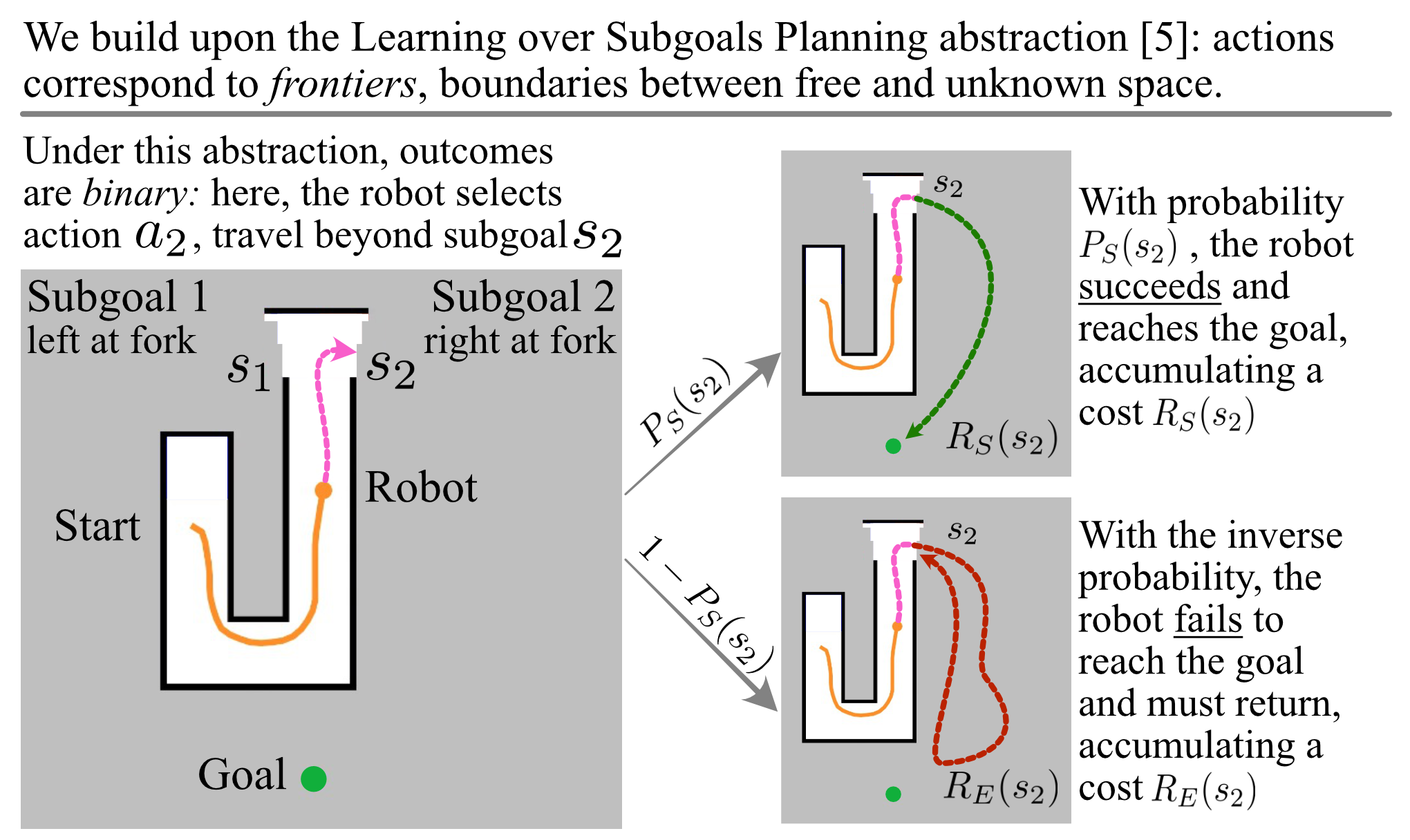}
    \vspace{-5pt}
    \caption{\textbf{Our robot's actions correspond to boundaries between free and unseen space.} The robot can leave observed space through either boundary: via subgoal $s_1$ or $s_2$. Upon selecting action $a_2$, the robot reaches the goal with probability $P_S$ and incurs an expected cost $R_S$, or is turned back (probability $1-P_S$), accumulates cost $R_E$ and selects another action.} \label{fig:lsp-example}
    \vspace{-10pt}
\end{figure}

\subsection{Preliminaries: Long-Horizon Model-based Navigation under Uncertainty via Learning over Subgoals Planning}
\label{sec:lsp}
Our robot relies on the learning-augmented model-based planning abstraction of Stein et al.~\cite{pmlr-v87-stein18a} for high-level navigation through partially-revealed environments.
Under this abstraction, high-level \emph{exploratory actions} correspond to \emph{frontiers}---boundaries between free and unseen space; each such action consists of navigating to a \emph{subgoal} (a point on the frontier) and then revealing the region of space beyond, often in an effort to reach the unseen goal. 

Consistent with the LSP action abstraction, planning under the LSP model is done over an abstract belief state: a tuple $b_t = \{m_t, q_t\}$, where $m_t$ is the current map of the environment, and $q_t$ is the robot pose.
Each high-level action $a_t \in \mathcal{A}(\{m_t, q_t\})$ has a binary outcome: with probability $P_S(a_t)$, the robot \emph{succeeds} in reaching the goal or (with the inverse probability $1 - P_S(a_t)$) fails to reach the goal.

Upon selecting an action $a_t$, the robot must first move through known space to the boundary, accumulating a cost $D(m_t, q_t, a_t)$.
If the robot succeeds in reaching the goal, it accumulates a \emph{success cost} $R_S(a_t)$, the expected cost for the robot to reach the goal, and no further navigation is necessary.
Otherwise, the robot accumulates an \emph{exploration cost} $R_E(a_t)$, the expected cost of revealing the region beyond the subgoal of interest and needing to turn back. The robot must subsequently choose another action $a_{t+1} \in A_{t+1} \equiv \mathcal{A}(\{m_t, q(a_t)\})\setminus \{ a_t \}$.
Fig.~\ref{fig:lsp-example} shows a schematic overview of the LSP state and action abstraction.

Determining the goodness of an exploratory action requires making predictions about what lies in unseen space.
So as to avoid the computational and practical challenges of exhaustive belief-space planning, the LSP abstraction relies on learning to estimate the statistics of unseen space for each exploratory action---the \emph{subgoal properties} $P_S$, $R_S$, and $R_E$.
We are interested in utilizing available sparse knowledge in the partial map to make quick relational inference about the unseen spaces for which the graph neural network~\cite{peter2018} shows promising performance.
So, we adopt a similar strategy as Arnob and Stein~\cite{arnob2023lspgnn} where a graph neural network variant with dynamic attention mechanism GATv2Conv~\cite{brody2021attentive} is used to estimate the subgoal properties.
The subgoal property estimator consumes a sparse graphical representation of the entire map, so that predictions about unseen space are informed by all of the robot's knowledge and experience thus far.

Under the LSP planning model, the expected cost of taking an action $a_t$ from belief state $b_t = \{ m_t, q_t\}$ is
\begin{equation}\label{eq:lsp-planning}
\begin{split}
    Q(&\{m_t, q_t\}, a_t\in  \mathcal{A}) = D(m_t, q_t, a_t) + P_S(a_t) R_S(a_t) \\
    & + (1-P_S(a_t)) \left[R_E(a_t) + \min_{a_{t+1}}Q(\{m_t, q(a_t)\},a_{t+1}) \right]
    \end{split}
\end{equation}
Planning via this abstraction is \emph{reliable by design} and \emph{complete}, as the robot is always traversing known space and making progress towards revealing unseen space.

It is thus the role of high-level planning via the LSP planning abstraction to select the exploratory action (and thus space to reveal) that will minimize expected cost. Once a high-level action is chosen, the robot selects the low-level primitive action that most makes progress towards that frontier through known space, updating the map as space is revealed and replanning when necessary; we use the notation $\pi^{\text{\tiny{}LSP}}_{\textit{\tiny{}mp}}(m_t))$ to mean the policy that returns the primitive action specified by LSP given the partial map $m_t$.

\subsection{Planning: Actively Seeking Valuable Information}
\label{sec:plan-with-value-of-info}

Under the LSP abstraction, temporally-extended \emph{exploratory actions} correspond to revealing a part of unseen space beyond its corresponding frontier. Under the LSP model, this begets only an exploration \emph{cost}, neglecting the possible benefits associated with revealing this region and the information it contains. It is a key insight of our approach that the cost associated with each exploratory action should be offset by the value of the information revealed during the exploration, $V_I(a_t)$, which represents how much performance would improve if the planner were provided an updated map with the target area revealed.

Thus our planning abstraction augments Learning over Subgoals planning Eq.~\eqref{eq:lsp-planning} to incentivize selection of actions that provide information valuable to improving planning performance:
\begin{equation}\label{eq:lsp-aig-planning}
\begin{split}
    Q(&\{m_t, q_t\}, a_t\in  \mathcal{A}) = D(m_t, q_t, a_t) + P_S(a_t) R_S(a_t) \\
    & + (1-P_S(a_t)) \Big[R_E(a_t) - V_I(a_t) \\ & + \min_{a_{t+1}}Q(\{m_t, q(a_t)\}, a_{t+1})\Big]
\end{split}
\end{equation}
Our Eq.~\eqref{eq:lsp-aig-planning} allows reliable navigation in partially-revealed environments and affords active information gathering to improve long-horizon behavior.
However, making use of Eq.~\eqref{eq:lsp-aig-planning} would require knowing the expected value of information associated with revealing each region of unseen space, an onerous task that in general requires significant computation and access to the underlying distribution over environments. Instead, we will estimate $V_I(a_t)$ via a learned model, leveraging a data generation procedure described in the next section.

\subsection{Determining the Value of Information of an Action}
\label{sec:compute-value-of-info}

The value of information $V_I$ associated with exploratory action $a_t$ is the cumulative difference in performance between LSP-driven plans that do and do not have access to the unseen space that would be revealed via execution of high-level action $a_t$ from the partial map $m_t$.
To compute this value at training time, we recognize that the value of information associated with action $a_t$ can be approximately computed as the sum of individual ``one step'' values of information $v_I$ along a trajectory, where $v_I$ is how much more progress towards the goal the robot could have made in one time step if the space corresponding to $a_t$ been revealed to it.

This \emph{one-step value of information} $v_I$ is defined via the policy $\pi^{\text{\tiny{}LSP}}_{\textit{\tiny{}mp}}$, which returns the primitive action that best makes progress towards the high-level action chosen by the high-level planner, as described in Sec.~\ref{sec:lsp}. The value $v_I$ corresponds to the difference in cost-to-go (progress towards the goal measured via the known map $m_{\text{\tiny{}known}}$) between the policy given only the partial map $m_t$ and the one in which the space corresponding to $a_t$ is revealed, notated as $m_t \cup a_t$. Formally, this is written as follows:
\begin{equation}\label{eq:lsp-aig-voi}
    v_I(m_t, a_t) = Q(m_{\text{\tiny{}known}}, \tikzmark{Qm}\pi^{\text{\tiny{}LSP}}_{\textit{\tiny{}mp}}(m_t)) -
    Q(m_{\text{\tiny{}known}}, \tikzmark{Qma}\pi^{\text{\tiny{}LSP}}_{\textit{\tiny{}mp}}(m_t \cup a_t))
\end{equation}
\includegraphics[width=.48\textwidth]{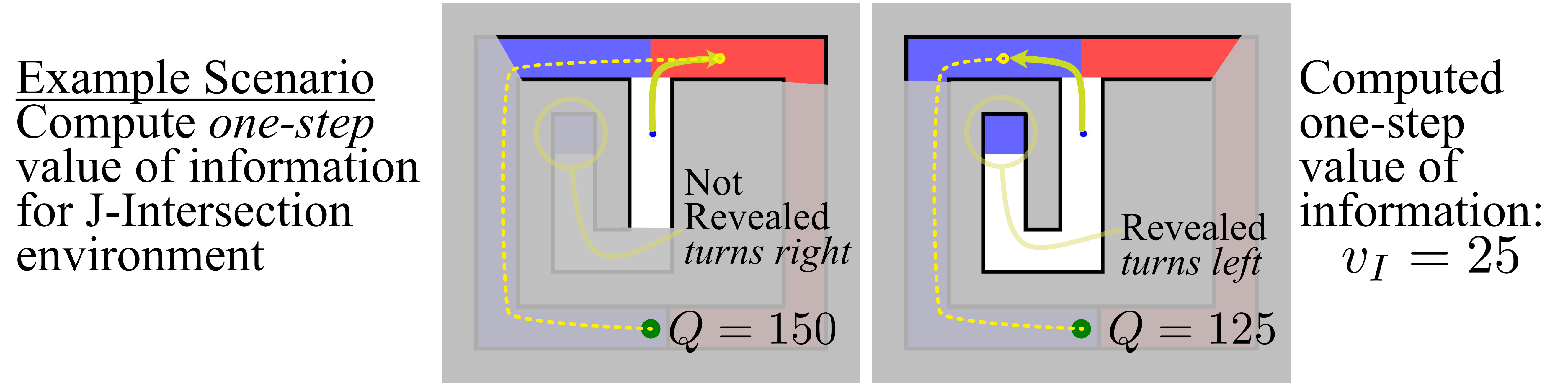}
\definecolor{light-gray}{HTML}{74729A}
\begin{tikzpicture}[remember picture,overlay]
        \draw[light-gray, line width=0.5pt,-] (pic cs:Qm) ++(-1.2, -0.15) -- ++(2.4, -0.0);
        \draw[light-gray, line width=0.8pt,->] (pic cs:Qm) ++(-0.0, -0.15) -- ++(0, -0.2) -- ++(0.2, 0) -- ++(0, -0.2);
        \draw[light-gray, line width=0.5pt,-] (pic cs:Qma) ++(-1.2, -0.15) -- ++(3.0, -0.0);
        \draw[light-gray, line width=0.8pt,->] (pic cs:Qma) ++(-0.0, -0.15) -- ++(0, -0.2) -- ++(-0.6, 0) -- ++(0, -0.2);
\end{tikzpicture}

\begin{figure}[t] 
  \includegraphics[width=.48\textwidth]{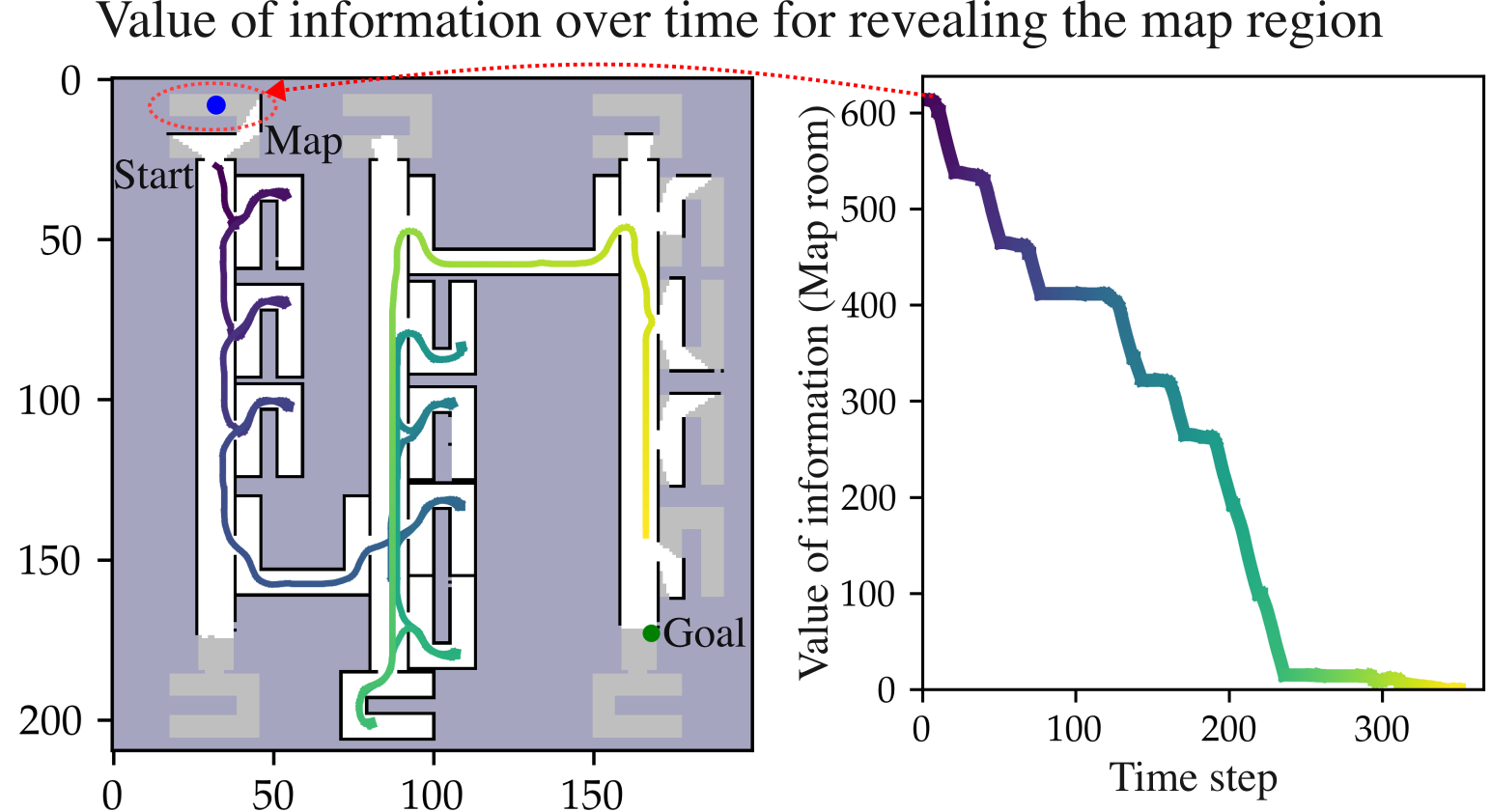}
  \vspace{-5pt}
  \caption{
  \textbf{Value of information training data example.}
  We show the total value of information $V_I$ (cumulatively summed over the one step value of information $v_I$) for the action that contains the map information. 
  Color signifies time step in both plots, enabling easy visual correspondence between the two.
  }
  \vspace{-10pt}  
  \label{fig:voi-calc}
\end{figure}

The total value of information $V_I$ for action $a_t$ is thus the sum of each one-step contribution $v_I$ for the remainder of travel. Training data is computed by deploying the base LSP policy, which plans via Eq.~\eqref{eq:lsp-planning}, and computing the one step value of information $v_I$ for all steps along the trajectory for each exploratory action that does not lead to the goal.
Fig.~\ref{fig:voi-calc} shows an example of the total computed $V_I$ over time for an action $a_t$ that contains the map room in our parallel hallway environment.

\subsection{Data Generation and Training}
\label{sec:gen-train-data}
To train our graph neural network, we require training data collected via offline navigation trials from which we can learn to estimate the subgoal properties ($P_S$, $R_S$, $R_E$, and $V_I$) for each subgoal node in the graph.
During an offline training phase, we conduct trials in which the robot navigates from start to goal and generates labeled data at each time step.
Training data consists of environment graphs $G$---with
input features and labels associated with each subgoal node.

Each graph node is given an observation---a node feature---from which the subgoal properties ($P_S$ , $R_S$ , $R_E$, and $V_I$) in Eq.~\ref{eq:lsp-planning} will be estimated via the graph neural network. 
Node features are 7-element vectors: (i) a 4-element one-hot semantic class (or color) at the location of the node, (ii) the number of neighbors of that node, (iii) a binary indicator of whether or not the node is a subgoal, and (iv) a binary indicator of whether the node is the goal node. 
We additionally include a single edge feature, associated with each edge in the graph: the geodesic distance between the nodes it connects. 
Owing to the presence of a goal node connected to every other node, the edge features provides each node its distance to the goal. 
See~\cite{arnob2023lspgnn} for more details. 

To compute the labels for our training data, we use the underlying known map $m_{\text{\tiny{}known}}$ to determine whether or not a path to the goal exists through a subgoal.
Using this information, we record a label for each subgoal that corresponds to a sample of the probability of success $P_S$ and from which we can learn to estimate $P_S$ using cross-entropy loss.
Labels for the other subgoal properties are computed similarly: labels for the success cost $R_S$ correspond to the travel distance through unknown space to reach the goal, for when the goal can be reached, and the exploration cost $R_E$ is a heuristic cost corresponding to how long it will take a robot to realize a region is a dead end, approximated as the round-trip travel time to reach the farthest reachable point in unseen space beyond the chosen frontier.
We compute the label value of information as described in Sec.~\ref{sec:compute-value-of-info}
This data and collection process mirrors that of LSP-GNN~\cite{arnob2023lspgnn}; readers are referred to their paper for additional details.

We repeat the data collection process for each step over hundreds of trials for each training environment.
So as to generate more diverse data, we switch between the known-space planner and an optimistic (non-learned) planner to guide navigation during data generation.
The details of each environment can be found in Sec.~\ref{sec:results}.

%% file: 6_results.tex
\section{Experimental Results}
\label{sec:results} 
We conduct simulated experiments in our \emph{J-Intersection} (Sec.~\ref{sec:example-case}, Sec.~\ref{sec:results:jint}), \emph{Parallel Office} (Sec.~\ref{sec:results:hallways}), and \emph{Ring Office} environments.
For each trial, we evaluate the performance of these planners:
\begin{LaTeXdescription}
\item[Non-Learned Baseline] Optimistically assumes the unseen space to be free and plans via grid-based A$^{\!*}$ search.
\item[LSP-GNN (learned baseline, \cite{arnob2023lspgnn})] Plans via Eq.~\eqref{eq:lsp-planning} using a graph neural network learning backend that consumes a sparse graphical representation of the entire (partially-revealed) map to estimate subgoal properties.
\item[LSP-AIG (Active Information Gathering, ours)] Plans via Eq.~\eqref{eq:lsp-aig-planning}, using an additional graph neural network to estimate the value of information for each exploratory action, and so seeks to reveal parts of the environment containing knowledge expected to improve planning performance.
\item[Map-Seeker (systematic baseline)] This approach, exclusive to the parallel hallway environment, seeks the map in the closer end of the top hallway first and then the other before it finds the map to head to the goal.
\end{LaTeXdescription}
For each planner, we compute the average navigation cost across many (at least 100) random maps from each environment.

\begin{table}[t]
    \begin{center}  
    \caption{Avg. Cost over Multiple Trials in all Tested Environments}\label{table:hallway-stats}
        \begin{tabular}{c|cccc}
            \toprule
            \textbf{Planners} & Naive & LSP-GNN & LSP-AIG & Map-\\
            \textbf{Environment} & Non-Learned & Learned & Ours & seeker\\
            \hline
            J-Intersection & $133.59$ & $107.82$ & \textbf{93.84} & -\\
            Ring-Office & $542.64$ & $310.57$ & \textbf{196.65} & - \\
            Parallel-Hallway & $252.35$ & $253.95$ & \textbf{228.11} & $257.99$ \\
            \bottomrule
        \end{tabular}
    \end{center}
    \vspace{-20px}
\end{table}

\subsection{J-Intersection Environment}
\label{sec:results:jint}
We first demonstrate that our LSP-AIG planner exhibits the information seeking behavior we expect in our motivating J-Intersection environment, as described in Sec.~\ref{sec:example-case}.
We conduct 100 trials for each planner in this environment and evaluate their performance. We report the average cost for each in Table~\ref{table:hallway-stats}.
Across all trials, our LSP-AIG planner \emph{always} correctly pursues the innermost part of the map and the information it contains; it subsequently makes use of to more quickly reach the unseen goal than is possible for either other planner, neither of which both reveal and make use of the hidden information. As such our LSP-AIG exhibits improvements of 36.9\% and 13.0\% over the non-learned baseline and LSP-GNN learned baseline, respectively.

\subsection{The Ring Office Environment}
\label{sec:results:ring-office}
\begin{figure}[t]
  \includegraphics[width=.48\textwidth]{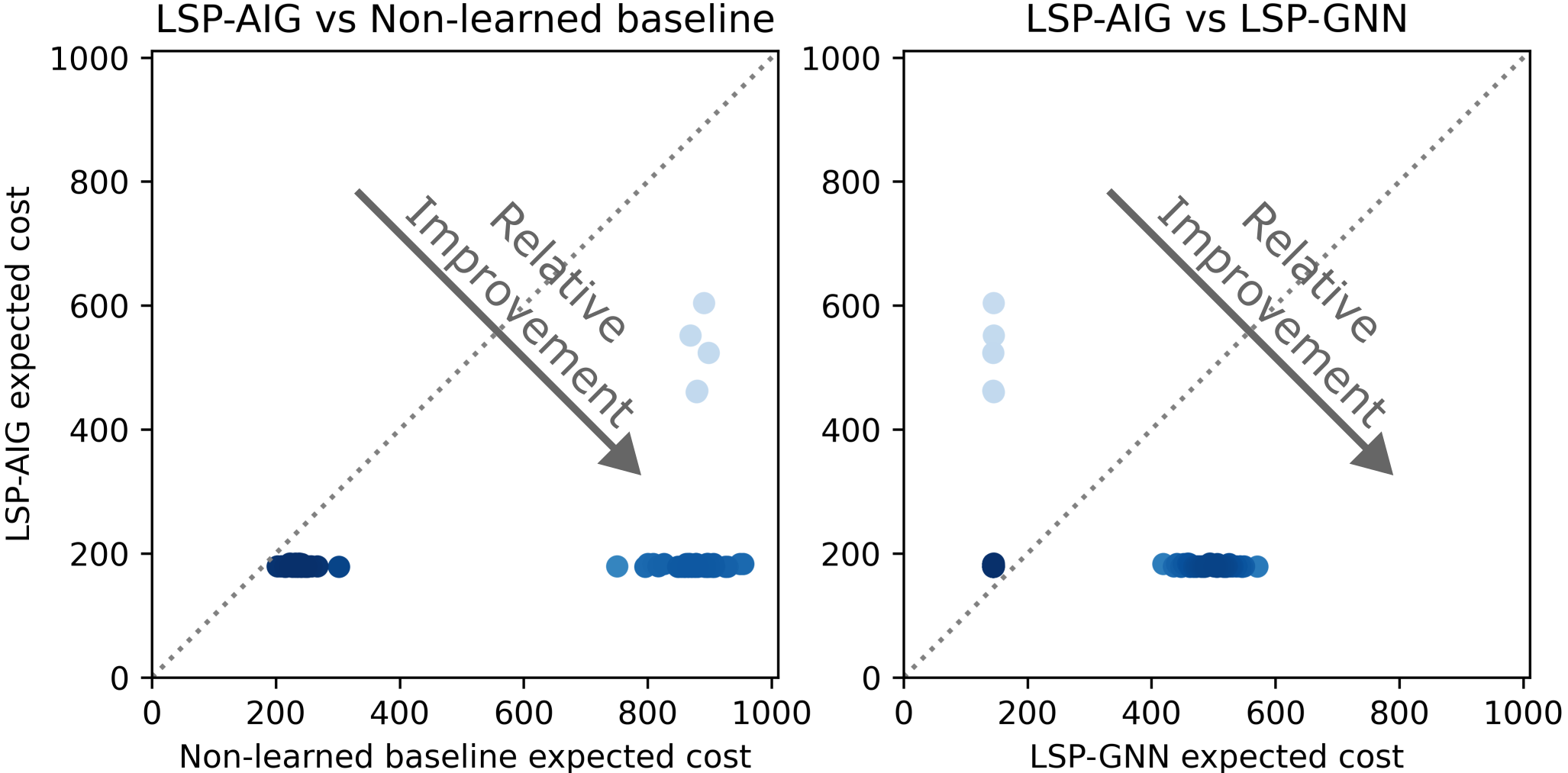}

  \vspace{5pt}
  
  \includegraphics[width=.48\textwidth]{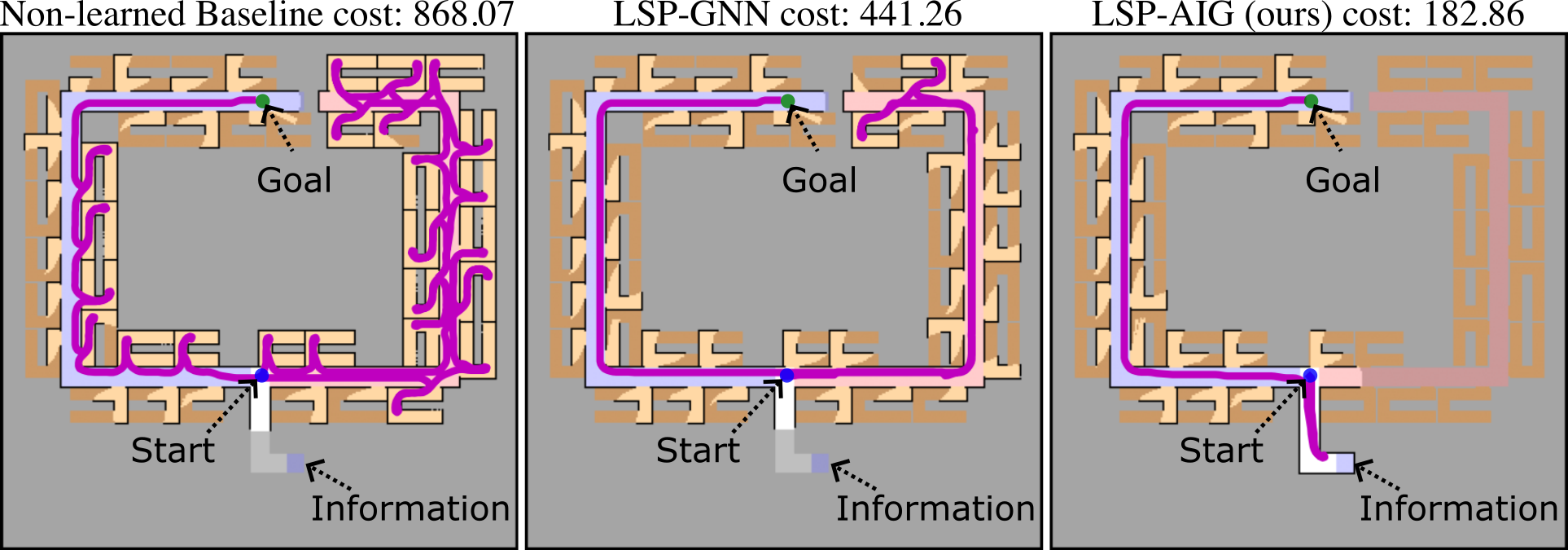}
  \vspace{-5pt}
  \caption{\textbf{Ring Office environment Results: scatter-plots and example trials.}
  Our LSP-AIG planner outperforms both the non-learned baseline and the LSP-GNN planners in 100 trials by actively gathering the information required to efficiently navigate in our Ring Office environment.}
  \vspace{-15pt}
  \label{fig:new-j-scatter}
\end{figure}
Our \emph{Ring-Office} environment (see Fig.~\ref{fig:new-j-scatter}) is, at a high-level, similarly structured to our \emph{J-Intersection} environment, in that, for both, an out-of-the-way region of unseen space contains information key to selecting the proper-color hallway to more quickly reach the goal. The \emph{Ring-Office} environment is both much larger than the \emph{J-Intersection} and additionally contains many dead-end rooms that flank the hallways, making the environment more challenging for all planning approaches. For all trials, the robot begins at the center of a three way intersection, from which it can choose to follow either hallway or to reveal the space that contains key information about which hallway is best.

We train the simulated robot on data from 200 maps and evaluate it in a separate set of 100 maps.
We show the average performance of each planning strategy in Table~\ref{table:hallway-stats} and include scatterplots of the relative performance of different planners for each trial in Fig.~\ref{fig:new-j-scatter}. 
The robot planning with our LSP-AIG approach knows to actively gather the information it needs to reduce its uncertainty about how best to reach the goal, and so achieves a 63.76\% improvement in average cost versus the optimistic Non-Learned baseline planner, and a 36.68\% improvement over the LSP-GNN baseline planner.

We highlight one trial in Fig.~\ref{fig:new-j-scatter}, in which the robot is tasked to navigate from the intersection to the goal.
From the intersection, the information about which hallway will lead to the goal is unavailable and the probability that either hallway leads to the goal is 50\% each.
The information is kept out of view from the intersection in the bottom right that our planner LSP-AIG actively seeks and then makes informed choice in the direction of the goal (left) from the intersection.
The non-learned baseline runs into many dead-end rooms, optimistically assuming anything it cannot see to be free space.
While the learned baseline LSP-GNN avoids dead-end rooms, it does not actively gather information and so selects the incorrect hallway.

\subsection{The Parallel Hallway Environment}\label{sec:results:hallways}

\begin{figure}[t]
  \includegraphics[width=.48\textwidth]{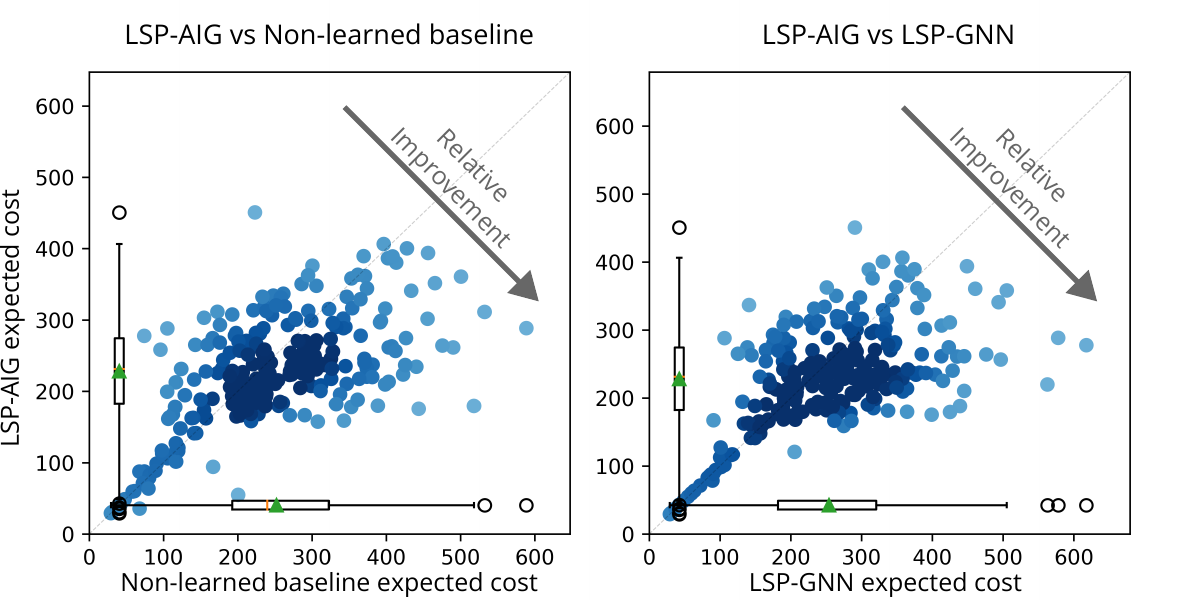}
  \includegraphics[width=.48\textwidth]{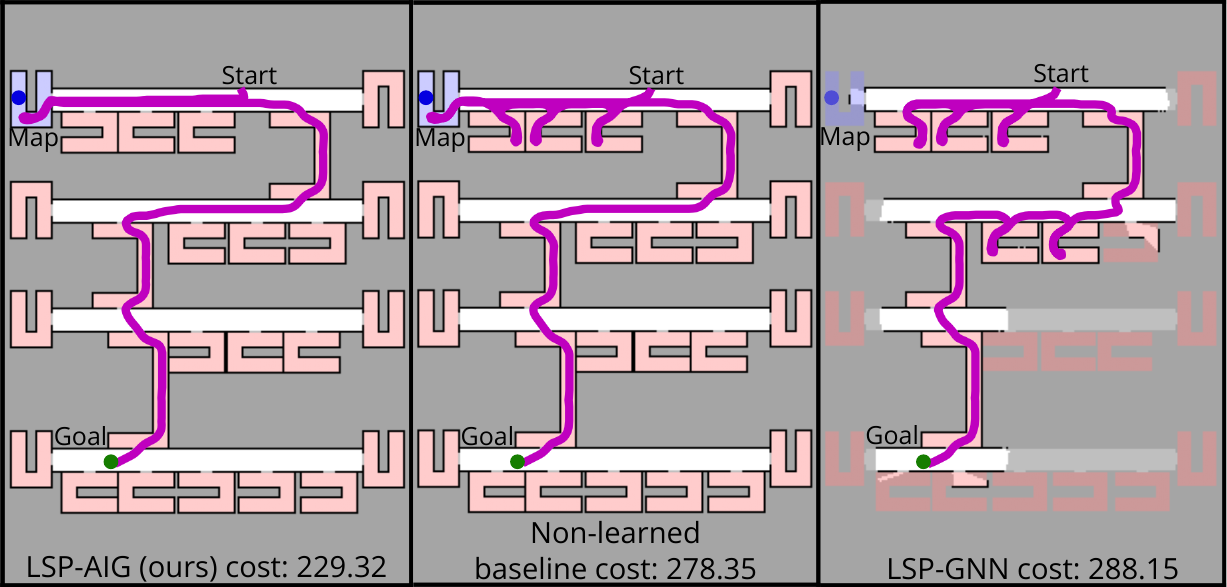}
  \vspace{-10pt}
  \caption{\textbf{Parallel Hallway Results: scatter-plots and example trials.} Our learning-informed LSP-AIG planner outperforms the non-learned baseline and the learned baseline LSP-GNN planners by actively gathering the map information required to  efficiently navigate in our Parallel Hallway environment.}
  \label{fig:scatter-plot-hallway}
\end{figure}

Our \emph{Parallel Hallway} environment (see Fig.~\ref{fig:scatter-plot-hallway}) consists of parallel hallways with rooms that border and connect them.
Our procedurally-generated maps contain three room types: (i) \emph{dead-end} rooms, (ii) \emph{passage} rooms that provide connections between neighboring parallel hallways, and (iii) a single \emph{map} room that contains a map of the entire environment.

Only one passage room exists between a pair of hallways, and so the robot must discover this room if it is to travel to a neighboring hallway.
Environments are generated such that the dead-end rooms and the passage rooms are of the same semantic ``color'' (red) distinct from that of the map room (blue).
The environment is such that traveling between hallways frequently requires a non-trivial amount of trial and error: as the passages and dead-ends are indistinguishable without entering them, the robot must enter many dead-end rooms before discovering the passage that allows it to make progress towards the goal, placed far away from the robot's starting location.

Critically, when the robot enters the map room the entire environment is revealed as well. While the map room will never contain the goal---and will routinely involve traveling away from the goal to find it---seeking out the map room when appropriate is a prime example of active information seeking behavior; once the map room (and thus the remainder of the environment) is revealed, the robot can make quick progress towards the goal without the need to seek out the passages between hallways. As shown in Fig.~\ref{fig:intro-explain}, good behavior in this environment involves first seeking out the map room, requiring that the robot understand the value of the information it provides.

We train the simulated robot on data from 500 distinct procedurally generated maps and evaluate it in a separate set of 250 distinct procedurally generated maps.
We show the average performance of each planning strategy in Table~\ref{table:hallway-stats} and include scatterplots of the relative performance of different planners for each trial in Fig.~\ref{fig:scatter-plot-hallway}.
The robot planning with our LSP-AIG approach achieves a 10.62\% improvement in average cost versus the optimistic Non-Learned Baseline planner, and a 11.33\% improvement over the base LSP-GNN planner, owing to our planner's ability to recognize the expected value of actively seeking out the map room early in navigation, going out of its way to improve performance overall.
Critically, though our planner is not told that there exists a map room nor what it's purpose is, it discovers during training that revealing this part of the environment will meaningfully improve planning performance and so actively gathers the information it provides during deployment. We show a representative trial in Fig.~\ref{fig:scatter-plot-hallway}.


\begin{figure}[t]
  \includegraphics[width=.48\textwidth]{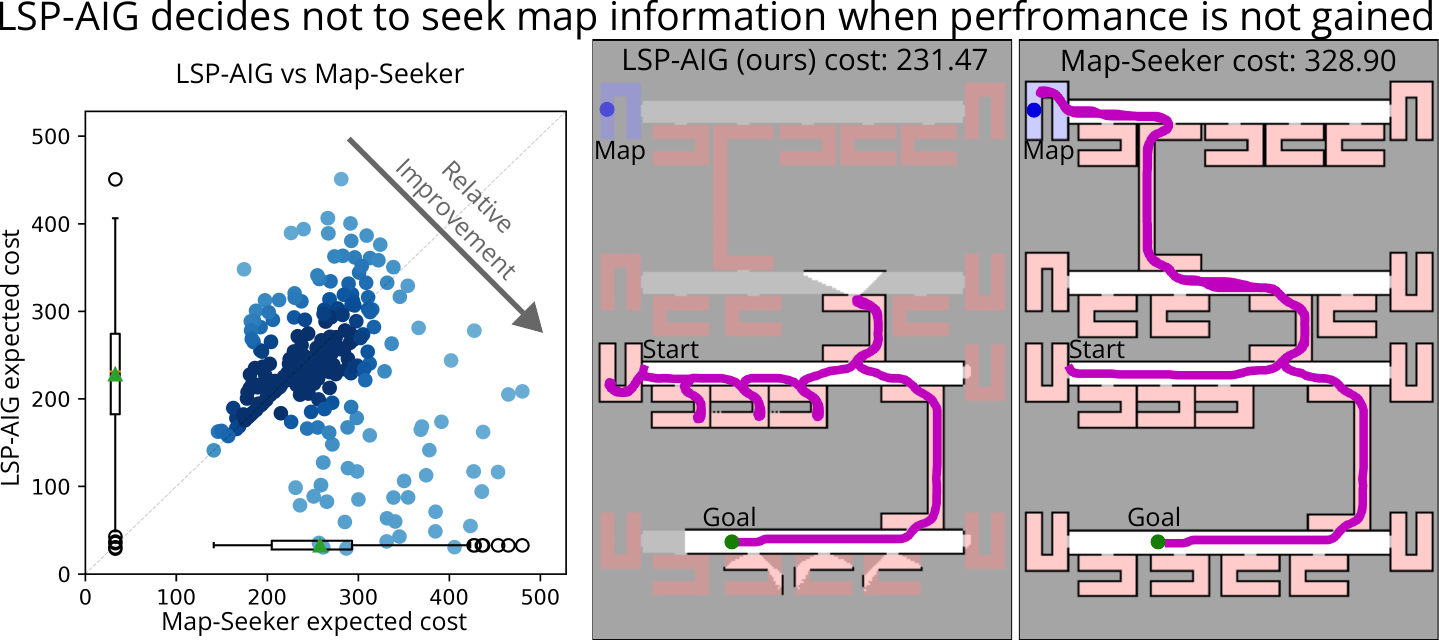}
  \vspace{-5pt}
  \caption{
  \textbf{Our LSP-AIG planner only seeks information when expected to be of sufficient value to improve performance.}
  We show one such trial in our \emph{Parallel Hallway} environment where our robot begins in the hallway next to the goal. The robot decides not to seek the map room, since it knows the expense would not justify the benefit.
  }
  \vspace{-10pt}
  \label{fig:middle-trajectory}
\end{figure}
We additionally note that our planner does not learn a policy that \emph{blindly} seeks the information the map room provides; instead, it seeks the map room when appropriate to improve its plan: when the value of information is sufficient to justify the expense of going out of its way to reveal it.
To illustrate this point, we conducted additional trials in which the robot is deployed in the hallway next to the goal, and so is both closer to the goal (so that there are fewer dead-end rooms to encounter, reducing the value of the information) and farther from the map room increasing the cost to reach it.
In Fig.~\ref{fig:middle-trajectory}, the robot is uncertain and heads in the direction of the map room, yet quickly realizes the effort is not worth the cost and turns back to seek the goal.

In summary, our results highlight the effectiveness of our planner, which has learned the value associated with revealing unseen regions of space, without any human-provided policy or guidance, and takes action to actively seek out valuable information when appropriate.

%% file: 8_conclusion_future_work.tex
\section{Conclusion and Future Work}\label{sec:conclusion-future}
We present a reliable model-based approach to navigation under uncertainty capable of actively gathering information to improve long-horizon performance.
Our planning approach uses a graph neural network to estimate the value of information of exploratory actions that reveal regions of unseen space and uses those estimates to encourage information seeking behavior when appropriate to improve performance. Our approach includes a process for efficiently generating the data necessary for training our planner during an offline training phase.
Substantiating our theoretical contributions, we demonstrate improved performance in simulated large-scale office-like environments.
In future work, we envision passing more complex sensory input to the robot, allowing it to estimate the goodness of exploratory actions using inputs from image sensors or semantically-segmented images.

%% file: 9_acknowledgement.tex
\section{Acknowledgement}\label{sec:ack}
We would like to thank Jana Kosecka, George Konidaris and Kevin Doherty for their thoughtful feedback on this work. 
This material is based upon work supported by the National Science Foundation under Grant No. 2232733.